\documentclass[letterpaper]{article}

\usepackage{natbib,alifeconf}  %% The order is important
\usepackage{amsmath, amssymb}
\usepackage{url,hyperref,cleveref}
\usepackage{booktabs}
\usepackage{graphicx} % Required for inserting images
\usepackage{subcaption}
\usepackage{tikz}
\usepackage{glossaries}
\glsdisablehyper
\usetikzlibrary{fit,positioning,backgrounds}
\usepackage[linewidth=0.75pt,ticklabelscale=0.75,labelscale=1,significancemarksscale=0.75]{plotmacros}
\usepackage[basecolor=gray,size=1mm,nOfSensorTypes=5]{vsrs}
\usepackage[inline,shortlabels]{enumitem}
\usepackage{cuted}
\usepackage{balance}

\colorlet{col.bo}{black}
\colorlet{col.rl}{blue}
\colorlet{col.left}{green}
\colorlet{col.right}{red}
\tikzset{
    linetype.darwin/.style = {dashed},
    linetype.lamarck/.style = {solid},
    linetype.sensor/.style = {dashdotted},
    linetype.nosensor/.style = {dotted},
    linetype.fixed/.style = {dotted}
}

\newacronym{rl}{RL}{reinforcement learning}
\newacronym{bo}{BO}{Bayesian optimization}
\newacronym{ea}{EA}{evolutionary algorithm}
\newacronym{er}{ER}{evolutionary robotics}
\newacronym{vsr}{VSR}{virtual soft robot}
\newacronym{ann}{ANN}{artificial neural network}

\makeatletter
\newcommand\blfootnote[1]{%
  \begingroup
  \renewcommand\thefootnote{}%
  \footnotetext[0]{#1}%
  \endgroup
}
\makeatother

\title{Lamarckian Inheritance in Dynamic Environments: How Key Variables Affect Evolutionary Dynamics}

\author{
    K. Ege de Bruin$^{1}$,
    Kyrre Glette$^{1,2}$, \and
    Kai Olav Ellefsen$^{1}$ \\
    \mbox{}\\
    $^1$Department of Informatics, University of Oslo, Norway \\
    $^2$RITMO, University of Oslo, Norway\\
    egedebruin@gmail.com
}

\begin{document}

\maketitle

\begin{strip}
    \centering
    \includegraphics[width=0.65\textwidth]{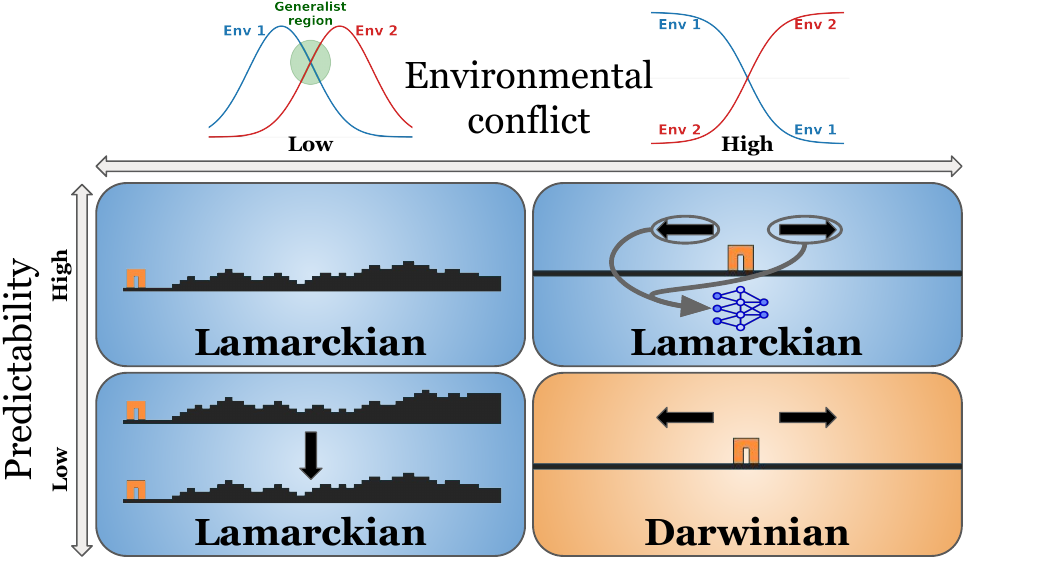}
    \captionsetup{hypcap=false}
    \captionof{figure}{We classify dynamic environments along two dimensions: environmental conflict (whether optimal control is conflicting in the environments) and predictability (how well agents can anticipate new environments). We show which inheritance mechanism, Darwinian or Lamarckian, is beneficial for each type of environmental change, together with example environments used in this work. The example environments are—shown counter-clockwise from the top left: a static rugged terrain, a dynamic rugged terrain, a bidirectional environment where the goal direction alternates every generation, and the same bidirectional environment with an added directional sensor for the agent.}
    \label{fig:teaser}
\end{strip}

\begin{abstract}
    The co-optimization of a robot’s body and brain presents a coupled challenge: the morphology constrains which control strategies are effective, while the control determines how well the morphology performs.
    To address this, we combine morphology optimization as evolution with controller optimization as lifetime learning, utilizing Lamarckian inheritance to transfer learned controller parameters from parent to offspring.
    In dynamic environments, existing literature presents conflicting evidence: while traditional evolutionary theory often suggests Lamarckian inheritance lacks benefit, recent studies in evolutionary robotics indicate it can improve performance.
    We hypothesize that this is because previous works have not included all relevant variables with dynamic environments.
    In this work, we show that the benefit of Lamarckian inheritance depends on two variables: how conflicting the environmental changes are to robot control, and the predictability of those changes for the robotic agent.
    Using virtual soft robots and two different learning approaches, Bayesian optimization and reinforcement learning, we show that Lamarckian inheritance only underperforms Darwinian inheritance when the changes are both conflicting and unpredictable.
    We find that adding a sensor to detect environmental changes restores the benefits for Lamarckian inheritance in conflicting environments, by allowing robotic agents to predict the need for a different behavior, thereby generalizing their control.
\end{abstract}

Submission type: \textbf{Full Paper}\\

Data/Code available at: \url{https://tinyurl.com/3z8v5nhs}
% Anonymous: https://tinyurl.com/2ewpswft
\blfootnote{
  \begingroup
  \renewcommand{\UrlFont}{\ttfamily} % ensures URLs are plain, not boxed
  \hypersetup{hidelinks}             % temporarily remove red box for this footnote
  \textcopyright  2026 K. Ege de Bruin, Kyrre Glette, Kai Olav Ellefsen. Published under a Creative Commons Attribution 4.0 International (CC BY 4.0) license.
  \endgroup
}

\section{Introduction}
In \gls{er}, \glspl{ea} are used to optimize robot or virtual creature design and control~\citep{Sims1994,Lipson2000,Faina2013,Nolfi2016,Cheney2016}.
The \gls{ea} needs to traverse a more complex search space due to the intertwined body and brain~\cite{mertan2025evolutionary}.
Consequently, there might be a mismatch between control and robot morphology, and a robot's morphology might be discarded because it has not been performing to its full potential due to poor control~\citep{Cheney2018,Luo2022,mertan2024investigating}. 

To deal with this problem, often morphology and controller optimization are separated from each other.
This can for example be done with an \emph{outer evolutionary loop}, to optimize morphologies, and an \emph{inner learning loop}, where every morphology goes through a controller optimization phase~\citep{Eiben2013}.
Adding a learning loop is an efficient way to find well-performing robots~\citep{Miras2020,Luo2022,Gupta2021,pigozzi2023morphology}.
This is because, in this case, new robot morphologies have more chance to reach more of their potential.
This does, however, come at the cost of requiring more resources to optimize a robot controller~\citep{Moreno2022,deBruin2025}.

If every new robot morphology goes through a controller-learning phase, one could follow a Darwinian approach, where control parameters are either randomly initialized for every robot morphology~\citep{Gupta2021}, or offspring robots inherit \emph{initial} control parameters from parents~\citep{Miras2020}.
However, this loses learned information new robots could use.
If a robot ends up with good control parameters \emph{after} learning, these learned parameters might be useful for other robots as well.
This is where a Lamarckian approach, where optimized control parameters are transferred from parent to offspring, could be used.
In previous work, it has been shown that Lamarckian inheritance~\citep{Jelisavcic2019,Luo2023,Harada2024} or other social learning approaches~\citep{Heinerman2015,deBruin2026} can improve performance.
These works are limited in that the environment is stationary over the evolutionary run.
Although recent studies still shows benefit of Lamarckian inheritance in dynamic environments~\citep{Luo2025,deBruin2026}, there is limited evidence and these studies are contradicted by other work~\citep{Paenke2007,Feldman1996,Ellefsen2013} that show no benefit of Lamarckian inheritance in dynamic environments. We hypothesize that these conflicting findings are a result of \emph{not including all the variables of relevance} to the study of when Lamarckian inheritance is beneficial.

In this work, we show that the benefit of Lamarckian inheritance over Darwinian inheritance in dynamic environments depends on the following two variables:
\begin{enumerate*}[(1)]
    \item how \emph{conflicting} the environmental changes are;
    and
    \item how \emph{predictable} the environmental changes are.
\end{enumerate*}
With controlled experiments with \glspl{vsr}, we demonstrate that these variables indeed determine when Lamarckian inheritance is beneficial in dynamic environments, and how sensing plays a role.

\section{Related work}
\subsection{Inheritance of learned traits in evolvable robots}
We have already mentioned that controller optimization, also known as learning, can improve performance in evolving robots~\citep{Miras2020,Luo2022,deBruin2025}. 
Controller optimization can be made more efficient by allowing optimized controller parameters to be exchanged between robots.
A common form of controller exchange is to inherit optimized controller parameters from parent to offspring, often described as Lamarckian inheritance.
\cite{Jelisavcic2019} showed that inheriting optimized controllers, rather than initial controllers, improves performance.
Similarly,~\cite{Luo2023} found that passing optimized parameters to offspring enhances adaptation.
Moreover,~\cite{Harada2024} use a transfer learning approach from parents to offspring where \gls{ann} weights are shared, and this improves performance over when weights are not shared.

Offspring robots can potentially receive information from other robots than their parents as well, for example in a social learning approach.
\cite{Heinerman2015} apply social learning to a population of morphologically identical robots, and show that adding social learning leads to better controllers faster, and that a combination of individual learning and social learning is beneficial.
\cite{deBruin2026} compare several social learning approaches, and show that in a population of morphologically similar robots it is best to have good teachers instead of teachers that are morphologically more similar.

\subsection{Dynamic environments}
Most of these works have shown benefits of information exchange in stationary environments, where the task stays the same and the environment does not have any changes.
However, dynamic environments can have an effect on the benefits of information exchange between robots.
Work within evolutionary theory suggests that the value of inherited information depends on environmental predictability.
If environmental conditions fluctuate rapidly, traits acquired or learned by parents may no longer be adaptive for offspring~\citep{Via1985}.
Similarly, studies from biology of nongenetic inheritance and transgenerational effects show that inherited environmental information only accelerates adaptation when parental environments provide information that can help predict the environmental state in  offspring~\citep{Jablonka2009,Bonduriansky2009}.

These insights from biology suggest that Lamarckian inheritance, like social learning or epigenetic effects, can lose its benefit in rapidly changing environments.
Several studies have also examined the effects of Lamarckian inheritance in computational models.
\cite{Paenke2007} show that Lamarckian inheritance loses its benefits over Darwinian inheritance in rapidly changing environments.
\cite{Feldman1996} show, in a different artifical life setting, again that information exchange, this time in the form of social learning, cannot succeed in an environment that changes every generation.
\cite{Sasaki1999} evolve \glspl{ann} in an environment where food can change between being poisonous or not, and they show that less information exchange not only showed more stable behaviour in dynamic environments, but also showed greater adaptability.
Finally,~\cite{Ellefsen2013} shows that environmental change within an individual's lifetime influences the costs and benefits of learning for an individual.
They show that both too stable and too variable environments will never select learning.
Together, these studies suggest that environmental variability strongly influences whether information exchange is beneficial.
Although most have been carried out in other areas of artificial life, they are directly relevant to \gls{er}, where a key goal is to develop robots capable of adapting to dynamic and unpredictable conditions.
Therefore, one might expect that we see a similar lack of benefit of Lamarckian inheritance in dynamic environments with evolving robots.

However, recent studies do show benefits of Lamarckian inheritance in dynamic environments for robots.
\cite{Luo2025} evolve robots where the environment changes from a flat to a rugged environment within several generations.
They do show a benefit of Lamarckian inheritance over Darwinian inheritance, with a quicker adaptation to environmental changes.
Finally,~\cite{deBruin2026} show that social learning still benefits evolving robots in dynamic environments.
The goal of this paper is to find reasons for this difference in observations in dynamic environments.

\section{Background}

\begin{figure}[t!]
    \centering

    \begin{subfigure}{\linewidth}
        \centering
        \includegraphics[width=\linewidth]{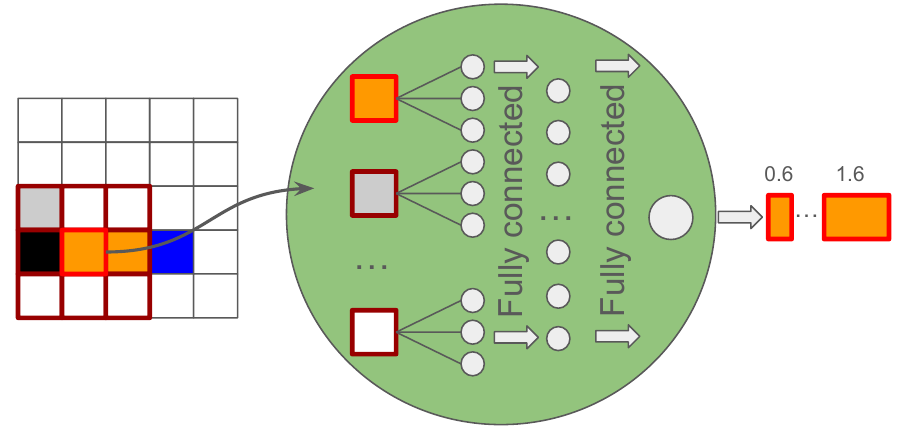}
        \caption{}
        \label{fig:brain}
    \end{subfigure}

    \vspace{1em}

    \begin{subfigure}{\linewidth}
        \centering
        \includegraphics[width=\linewidth]{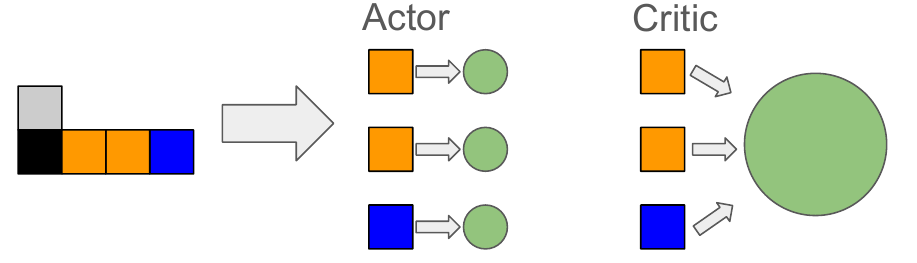}
        \caption{}
        \label{fig:actor-critic}
    \end{subfigure}

    \caption{
        Overview of the robot brain and its role in \gls{rl}.
        (a) Structure of the neural controller for a single voxel in an example robot. 
        Each voxel is controlled by an identical modular network, with shared weights across all voxel controllers. 
        The inputs consist of the voxel’s local state and information from its neighboring voxels. 
        The network outputs the control signal for that voxel. 
        (b) Actor-critic architecture used in the \gls{rl} setting. 
        The actor network has the same structure as the brain used for \gls{bo}. 
        The critic is a single centralized network that receives as input the states of all voxels and outputs a single scalar value estimate. 
        Note that the critic additionally receives the actions of all voxels as input.
    }
    \label{fig:combined}
\end{figure}

\subsection{\Acrfullpl{vsr}}
We use Evolution Gym (EvoGym)~\citep{Bhatia2021} as the simulator for our experiments.
EvoGym is a 2D \gls{vsr} (also named voxel-based soft robots) simulator.
There are four types of voxels: a rigid non-actuated voxel (\vsrevogym[2mm]{1}{1}{1}) that continuously stays in the same shape, a soft non-actuated voxel (\vsrevogym[2mm]{1}{1}{2}) that can change shape depending on the pushing or pulling from neighbouring voxels, an actuated voxel that can change its shape horizontally (\vsrevogym[2mm]{1}{1}{3}), and an actuated voxel that can change its shape vertically (\vsrevogym[2mm]{1}{1}{4}).
All voxels can be connected to each other orthogonally to create an entity that interacts with an environment.
The actuated voxels (\vsrevogym[2mm]{1}{1}{3} and \vsrevogym[2mm]{1}{1}{4}) are expanded or contracted by a value $u$, which is within the range $[0.6, 1.6]$, and corresponds to $u$ times its rest length.

The brain of the VSR is in charge of determining the target deformation.
We use a modular \gls{ann}–based controller, following the same architecture as in~\citep{Mertan2023}, since such controllers have been shown to work well for brain–body co-optimization.
Each voxel is controlled by the same \gls{ann}, but with different inputs depending on the voxel.
The \gls{ann} takes as input the horizontal and vertical velocity, and the volume of all voxels within a Moore neighbourhood of 1, including their own. 
Input signals for non-existant voxels are $0$.
We add a periodic time signal based on the simulation step modulo $25$, i.e., $c = \text{current-time} \bmod 25$, which produces a repeating sequence over 25 steps. 
This value is mapped linearly to an angle $\theta = \frac{2\pi}{25} c$, and the two inputs $\sin(\theta)$ and $\cos(\theta)$ are used as the cyclic time representation.
Finally, each voxel has a single touch sensor that is activated if the voxel is touching the ground.
This results in a total of $30$ input signals.
The hidden layer consists of $10$ neurons with ReLU activation functions, and there is a single output neuron with a sigmoid activation function.
We scale the output to fit the action range of $[0.6, 1.6]$.
We remark that, being composed of identical modules, this kind of brain is inherently transferable to any body (though it will likely not perform well from scratch), as the architecture of the network is independent from the shape of the body.
An overview of this network for the brain can be seen in Figure \ref{fig:brain}.

\subsection{\Acrfull{bo}}
\gls{bo} is an iterative numerical optimization technique based on a surrogate model of the problem to be solved.
It is generally more \emph{explorative} than other typical controller optimization methods, which means it often quickly finds good control parameters, but it often lacks in exploiting these parameters well.

Let $f: \mathbb{R}^p \to \mathbb{R}$ be a black-box function, \emph{i.e.}, a function which can be applied (\emph{observed}), but whose concrete structure is unknown.
The goal is to solve the maximization problem $\arg\max_{\vec{x} \in \mathbb{R}^p} f(\vec{x})$.

\gls{bo} proceeds by iteratively selecting evaluation points based on a surrogate model.
Let $\theta \in \Theta \subset \mathbb{R}^d$ denote the control parameters of a robot. 
\gls{bo} maintains a surrogate model $\hat{f}$ that provides a prediction of the performance given $\theta$.
This surrogate results in a predicted mean $\mu(\theta)$ and uncertainty $\sigma(\theta)$.
An acquisition function $a$ uses $\mu(\theta)$ and $\sigma(\theta)$ to balance exploration and exploitation and selects the next candidate point
$\bar{\theta} = \arg\max_{\theta \in \Theta} a(\theta; \hat{f})$.
The selected parameters are then evaluated on the robot, resulting in an objective value $f(\theta)$.
This new observation is added to the dataset and used to update the surrogate model $\hat{f}$.
The \gls{bo} then iteratively repeats the process of 
\begin{enumerate*}[(i)]
    \item selecting promosing parameters (\emph{candidate} parameters) $\bar{\vec{\theta}}$ using the acquisition function $a$,
    \item evaluating the candidate parameters on the robot resulting in an objective value $f(\bar{\vec{\theta}})$, 
    and
    \item updating the surrogate model $\hat{f}$ using the set $\{\bar{\vec{\theta}}, f(\bar{\vec{\theta}})\}$ as a \emph{sample}.
\end{enumerate*}

There are several options for the various components of \gls{bo}.
In this work, we use the Matern 5/2 kernel with a length scale of $10$ as the surrogate function $\hat{f}$ and the upper confidence bound as the acquisition function, with an exploration variable of $3$, which we optimize using the L-BFGS-B algorithm~\citep{Zhu1997}.
These settings have been used and worked well for optimizing the controller of robots for directed locomotion~\citep{Lan2021,VanDiggelen2024,deBruin2025,deBruin2026}.

\subsection{\Acrfull{rl}}
We also use \gls{rl} as a controller optimization method, which is more \emph{exploitative} than \gls{bo}, meaning it can find better control parameters if good ones are already found.
An \gls{rl} agent observes the environment at every timestep through its sensors and, according to its policy, maps the current observation (state) to an action.
In general this is similar to how the \gls{bo}-based controller functions: both methods define a controller (or policy) that determines the robot’s behavior.
The key difference is how this controller is updated.
In contrast to \gls{bo}, where the controller parameters remain fixed during an evaluation, an \gls{rl} controller is updated continuously while an episode is running.
Concretely, after every fixed number of timesteps (in our case, every fifth timestep), the policy is updated to maximize the expected cumulative reward.
The reward function depends on the task, and can be calculated similarly as the objective function.
For our simple locomotion tasks, both the objective value and the reward are computed from the distance traveled by the robot over time.

To optimize the policy, we apply the DDPG algorithm~\citep{Lillicrap2019}, which is an actor–critic method based on two \glspl{ann}.
The first network, referred to as the policy (or actor) network, represents the controller of the robot.
This network is identical to the previously described robot brain and consists of a modular \gls{ann}–based controller that outputs actions for the robot’s actuators given the current sensory state.
The second network is the critic network that estimates the value of a state–action pair, commonly denoted as the Q-value.
This value represents the expected cumulative future reward obtained by doing a given action in a given state and following the current policy.
The critic thus provides a learning signal that indicates how good a particular action choice is, which is then used to update the policy network.
Unlike the policy network, which is modular, the critic is implemented as a single, centralized network.
The reason for this difference is to give the critic an idea about the morphology of the robot so that it estimates the value function better.
Moreover, this lessens the number of parameters that are inherited from parent to offspring.
It takes as input the states \emph{and actions} of all actuated voxels of the robot and outputs a value estimating the quality of the corresponding state–action pair.
In our work, between the input and output layer are two hidden layers, consisting of respectively 128 and 64 neurons with ReLU activation functions.
The critic network is used to update the policy by providing a gradient-based learning signal.
For each observed state, the policy network outputs an action, which is evaluated by the critic through its estimated value. 
The policy parameters are then updated via backpropagation to maximize this estimated value, using the gradient of the critic with respect to the action.
An overview of the actor-critic network can be seen in Figure \ref{fig:actor-critic}.

\section{\Acrfull{ea}}
\subsection{Body evolution}
In our work, the genotype of a morphology is a $5x5$ grid where voxels are directly encoded in the position of the grid to create a robot.
In EvoGym, such a direct encoding is more often chosen~\citep{Mertan2023} due to its better performance in prior studies~\citep{Bhatia2021}.
For the initial robots, we iteratively add $10$ to $20$ random types of voxels out of the set $V=\{\vsrevogym[2mm]{1}{1}{1}, \vsrevogym[2mm]{1}{1}{2}, \vsrevogym[2mm]{1}{1}{3}, \vsrevogym[2mm]{1}{1}{4}\}$ on the $5x5$ grid to create an orthogonally connected robot.
To ensure orthogonally connected voxels, voxels need to be orthogonally connected to at least one existing voxel.
Mutations can either add, remove or change up to three voxels.
The size of the robot stays between $5$ and $25$ voxels, and if mutation causes a robot go out of this range, we try another mutation.

Morphology optimization is done using a simple \gls{ea}.
The population size is $100$, and we create $100$ offspring by mutating selected morphologies.
We use tournament selection for parent selection with a tournament size of $4$.
Because we are changing the environment over generations, the offspring replaces the current population.

\begin{figure}
    \centering
    \includegraphics[width=1\linewidth]{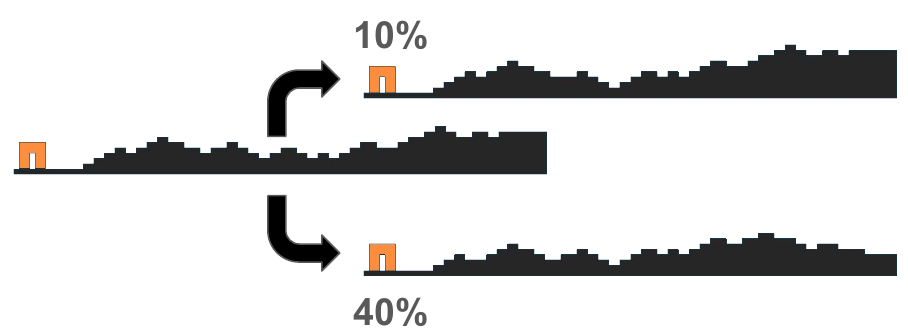}
    \caption{
        Example of a rugged environment. Percentage shows the difference to the original environment.
    }
    \label{fig:rugged-example}
\end{figure}

\subsection{Brain learning}
For the first generation of robots, we generate initial values by randomly sampling in $[-1,1]^{321}$, after which the controllers will be optimized using a learning loop.
For learning the brain, we use the two previously explained algorithms, \gls{bo} and \gls{rl}, and apply Darwinian and Lamarckian inheritance to them.
The two of them are chosen due to their different approaches in optimization, where \gls{bo} is more explorative and \gls{rl} more exploitative.
In our experiments we will show that they both have their benefits.

\subsubsection{Lamarckian inheritance}
To apply Lamarckian inheritance to \gls{bo}, we transfer samples from parent robots to offspring robots~\cite{deBruin2025,deBruin2026}.
For every new robot, the initial candidate brains to be evaluated are taken from the $8$ best samples from its parent.
We choose the best samples instead of the last samples, because \gls{bo} is inheritently explorative and latter samples can likely have a poor performance.
This means that every robot will re-evaluate the best samples of its parent.
After inheritance, the robots go through the \gls{bo} process with total of $50$ samples.

For \gls{rl}, the process is slightly different.
Firstly, the controller parameters are continously updated during episodes.
Therefore, only $1$ set of controller parameters are inherited from a robot's parent.
The controller parameters to be evaluated are the initial parameters of the best episode of its parent.
One could choose to inherit final parameters when using \gls{rl}, but we inherited the best to keep differences minimal with the \gls{bo} implementation.
Earlier experiments showed minimal differences in performance between inheriting the best or last parameters.
The robots will then go through the \gls{rl} process.
Note that the structure of the critic network of the \gls{rl} agent is morphology dependent, as illustrated in Figure \ref{fig:actor-critic}.
Therefore, the critic network parameters are not inherited by offspring robots, and are initialized randomly for every robot.
The number of episodes and timesteps per episode is equal to the number of samples and timesteps per sample in the \gls{bo} alternative, namely $50$ and $500$.

For both methods, the sample/episode with the best objective value will be used for the fitness of the robot body.

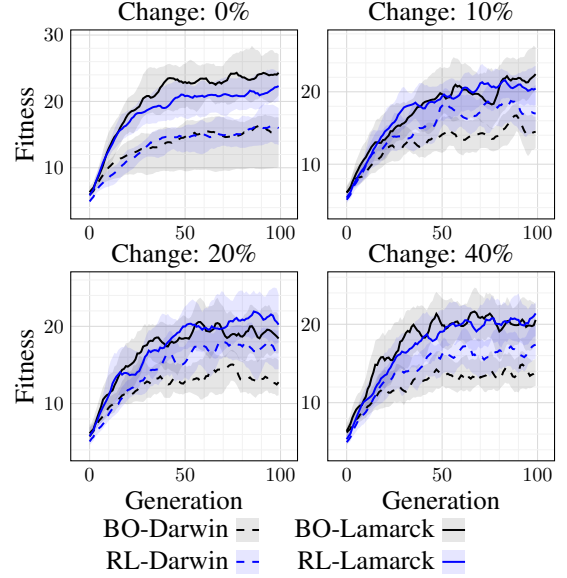
\begin{figure}
    \centering
    \begin{tikzpicture}
    \begin{groupplot}[
        group style={
            group size=2 by 2,
            xlabels at=edge bottom,
            ylabels at=edge left,
            horizontal sep=4mm,
            vertical sep=7mm
        },
        scale only axis,
        width=30mm,
        height=25mm,
        gridded,
        noinnerticks,
        xlabel={Generation},
        ylabel={Fitness}
    ]

    \nextgroupplot[title={Change: 0\%}]
    
        \lineminmax[lcolor=col.bo,ltype=linetype.darwin]{results/rugged/0/BODarwinian.txt}{}{x}{y}{ymin}{ymax}
        \lineminmax[lcolor=col.bo,ltype=linetype.lamarck]{results/rugged/0/BOLamarckian.txt}{}{x}{y}{ymin}{ymax}
        \lineminmax[lcolor=col.rl,ltype=linetype.darwin]{results/rugged/0/RLDarwinian.txt}{}{x}{y}{ymin}{ymax} 
        \lineminmax[lcolor=col.rl,ltype=linetype.lamarck]{results/rugged/0/RLLamarckian.txt}{}{x}{y}{ymin}{ymax} 
    
    \nextgroupplot[title={Change: 10\%}]
    
        \lineminmax[lcolor=col.bo,ltype=linetype.darwin]{results/rugged/10/BODarwinian.txt}{}{x}{y}{ymin}{ymax}
        \lineminmax[lcolor=col.bo,ltype=linetype.lamarck]{results/rugged/10/BOLamarckian.txt}{}{x}{y}{ymin}{ymax}
        \lineminmax[lcolor=col.rl,ltype=linetype.darwin]{results/rugged/10/RLDarwinian.txt}{}{x}{y}{ymin}{ymax} 
        \lineminmax[lcolor=col.rl,ltype=linetype.lamarck]{results/rugged/10/RLLamarckian.txt}{}{x}{y}{ymin}{ymax} 

    \nextgroupplot[title={Change: 20\%}]
    
        \lineminmax[lcolor=col.bo,ltype=linetype.darwin]{results/rugged/20/BODarwinian.txt}{}{x}{y}{ymin}{ymax}
        \lineminmax[lcolor=col.bo,ltype=linetype.lamarck]{results/rugged/20/BOLamarckian.txt}{}{x}{y}{ymin}{ymax}
        \lineminmax[lcolor=col.rl,ltype=linetype.darwin]{results/rugged/20/RLDarwinian.txt}{}{x}{y}{ymin}{ymax} 
        \lineminmax[lcolor=col.rl,ltype=linetype.lamarck]{results/rugged/20/RLLamarckian.txt}{}{x}{y}{ymin}{ymax} 

    \nextgroupplot[title={Change: 40\%}]
    
        \lineminmax[lcolor=col.bo,ltype=linetype.darwin]{results/rugged/40/BODarwinian.txt}{}{x}{y}{ymin}{ymax}
        \lineminmax[lcolor=col.bo,ltype=linetype.lamarck]{results/rugged/40/BOLamarckian.txt}{}{x}{y}{ymin}{ymax}
        \lineminmax[lcolor=col.rl,ltype=linetype.darwin]{results/rugged/40/RLDarwinian.txt}{}{x}{y}{ymin}{ymax} 
        \lineminmax[lcolor=col.rl,ltype=linetype.lamarck]{results/rugged/40/RLLamarckian.txt}{}{x}{y}{ymin}{ymax} 
    
    \end{groupplot}
    \end{tikzpicture}
    \begin{tabular}{rr}
        BO-Darwin \addlegendimageintext{shaded legend image={col.bo}{linetype.darwin}} &
        BO-Lamarck \addlegendimageintext{shaded legend image={col.bo}{linetype.lamarck}} \\
        RL-Darwin \addlegendimageintext{shaded legend image={col.rl}{linetype.darwin}} &
        RL-Lamarck \addlegendimageintext{shaded legend image={col.rl}{linetype.lamarck}}
    \end{tabular}
    \caption{
        Mean performance of the population over generations in a random rugged environment. Shaded areas indicate interquartile range (25th to 75th percentile). The lines are smoothened with a window size of 4 generations. The initial environment is set randomly, the percentages above the plots indicate how much the environments differ between subsequent generations. 
    }
    \label{fig:rugged}
\end{figure}

\subsubsection{Darwinian inheritance}
We compare both Lamarckian variants (\gls{bo}-based and \gls{rl}-based) to Darwinian variants, where there is no inheritance of learned behaviour.
Instead, offspring inherit initial control (before learning) parameters that are stored in the genotype of the parent.

In detail, we set the genotype space of the brain as $\mathbb{R}^{321}$.
For the first generation, this works the same as the control parameters for the first generation, \emph{i.e.}, the genotype values are randomly initialized.
Offspring robots then inherit the genotype values from their parent instead of the \emph{learned} control parameters.
We do apply mutation to the inherited genotype values, we do mutation by applying a perturbation sampled from the multivariate normal distribution $N(0, \sigma_\text{mut} \vec{I})$, $\vec{I} \in \mathbb{R}^{321 \times 321}$ being the identity matrix and $\sigma_\text{mut}=0.1$, \emph{i.e.}, we use the Gaussian mutation.

\section{Experimental analysis}

\begin{figure}
    \centering
    \includegraphics[width=1\linewidth]{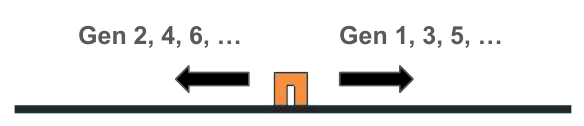}
    \caption{
        The bidirectional environment. It is a flat environment where the goal direction switches every generation.
    }
    \label{fig:bidirectional-example}
\end{figure}

We compare the four approaches, a Darwinian and Lamarckian variant of both \gls{bo} and \gls{rl}, in three settings.
The settings are chosen to investigate how the benefit of Lamarckian inheritance is dependent on the two relevant variables mentioned in the introduction.
After showing a case that confirms previous work in robotics that Lamarckian inheritance can still be beneficial in dynamic environments, we then show the impact of confliciting environmental change, and finally, the ability to predict changes between generations.
The task for all settings is locomotion, where the objective is to maximize distance along the $x$-axis in $500$ time steps.
We compute the distance by considering the $x$-coordinate of the center of mass of the robot at the beginning ($k=0$) and at the end ($k=500$) of the episode.
In every experiment, the total number of generations is $100$ and we execute for $20$ runs for every setting.

The settings used are chosen to show the importance of the two variables shown in Figure \ref{fig:teaser}, \emph{i.e.}, how conflicting the environmental changes are and the predictability for the agent.
In the first experiment, the environment is randomly rugged, where we compare different degrees of random changes between generations.
This is a \emph{non-conflicting} change, since a robot can generalize its movement to walk on any rugged terrain, and \emph{non-predictable} change for the agent, since it has no indication of the environmental changes.
We also include a stationary rugged environment, which we cast as a \emph{predictable} environment since it doesn't change.
In the following two settings, the robot is put on a flat surface, where the goal direction changes every generation, \emph{i.e.}, in one generation the robot's task is to move to the left and in the next generation the goal direction switches.
This is a \emph{conficting} environment, because movement in one direction causes a negative reward in the other direction.
The difference between the two settings is that in the last experiment, the robot has a sensor guiding it in the correct direction, making it a \emph{predictable} change for the robot.

\subsection{Rugged environment} \label{sec:exp-rugged}
\label{sec:exp-bidirectional}

The first dynamic environment we experiment with is a rugged environment with varying degrees of change between generations.
Environments in the initial generation are randomly generated subject to a set of constraints.
Each environment is a height profile built step by step over 100 positions, starting with a flat section where the height is fixed at 1 for the first 10 positions.
After that, the height can stay the same, go up by 1, or go down by 1, but upward or downward moves are only allowed if the previous step was flat.
The height is bounded between 1 and 10.
In subsequent generations, the environment changes according to a change percentage $c$. At each position, the height difference from the previous generation is preserved with probability $1-c$.
Otherwise, a new height change is sampled instead.
We ensure it remains valid under the previously mentioned environment constraints.
An example environment, and how the environment changes, is shown in Figure \ref{fig:rugged-example}.

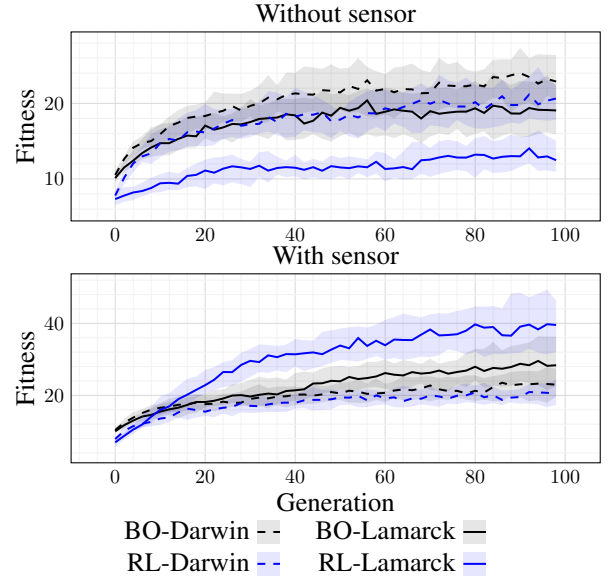
\begin{figure}
    \centering
    \begin{tikzpicture}[baseline=(current bounding box.center)]
        \begin{groupplot}[
            group style={
                group size=1 by 2,
                xlabels at=edge bottom,
                ylabels at=edge left,
                horizontal sep=4mm,
                vertical sep=7mm
            },
            scale only axis,
            width=70mm,
            height=25mm,
            gridded,
            noinnerticks,
            xlabel={Generation},
            ylabel={Fitness},
        ]
        \nextgroupplot[title={Without sensor}]
            \lineminmax[lcolor=col.bo,ltype=linetype.darwin]{results/conflicting/BODarwinian.txt}{}{x}{y}{ymin}{ymax}
            \lineminmax[lcolor=col.bo,ltype=linetype.lamarck]{results/conflicting/BOLamarckian.txt}{}{x}{y}{ymin}{ymax}
            \lineminmax[lcolor=col.rl,ltype=linetype.darwin]{results/conflicting/RLDarwinian.txt}{}{x}{y}{ymin}{ymax} 
            \lineminmax[lcolor=col.rl,ltype=linetype.lamarck]{results/conflicting/RLLamarckian.txt}{}{x}{y}{ymin}{ymax} 
        \nextgroupplot[title={With sensor}]
            \lineminmax[lcolor=col.bo,ltype=linetype.darwin]{results/sensing/BODarwinian.txt}{}{x}{y}{ymin}{ymax}
            \lineminmax[lcolor=col.bo,ltype=linetype.lamarck]{results/sensing/BOLamarckian.txt}{}{x}{y}{ymin}{ymax}
            \lineminmax[lcolor=col.rl,ltype=linetype.darwin]{results/sensing/RLDarwinian.txt}{}{x}{y}{ymin}{ymax} 
            \lineminmax[lcolor=col.rl,ltype=linetype.lamarck]{results/sensing/RLLamarckian.txt}{}{x}{y}{ymin}{ymax}
        \end{groupplot}      
    \end{tikzpicture}    
    \begin{tabular}{rr}
        BO-Darwin \addlegendimageintext{shaded legend image={col.bo}{linetype.darwin}} &
        BO-Lamarck \addlegendimageintext{shaded legend image={col.bo}{linetype.lamarck}} \\
        RL-Darwin \addlegendimageintext{shaded legend image={col.rl}{linetype.darwin}} &
        RL-Lamarck \addlegendimageintext{shaded legend image={col.rl}{linetype.lamarck}}
    \end{tabular}\
    \caption{
        Mean performance of the population over generations in a bidirectional environment. The top figure is for the setting without direction sensor, the bottom figure is for the setting with an added direction sensor to the robot brain. Every generation the goal direction switches. Shaded areas indicate interquartile range (25th to 75th percentile). The lines are smoothened with a window size of 2 generations.
    }
    \label{fig:bidirectional}
\end{figure}

We show the results of four values for $c$ in Figure \ref{fig:rugged}. Note that for $c=0\%$ there is no environmental change between generations.
These results confirm previous work in comparing Lamarckian to Darwinian inheritance in dynamic environments with evolving robots.
Namely, no matter the degree of change in the environment, there is still a benefit of Lamarckian inheritance.
One might notice that the benefit of Lamarckian inheritance is greater with no change of environment than some degree of change, but in any setting of $c$ there is still a clear benefit of Lamarckian inheritance.
Our hypothesis for this is that the environmental change is not conflicting enough to require a control change.
Therefore, learned control in one rugged environment can still be useful in a very different, but still rugged environment.

\subsection{Conflicting environment change}

Now that we have shown a case where the environment in not conflicting, and Lamarckian inheritance does indeed still show beneficial, we will show a case where the environment is conflicting.
The robot is now put on a flat plane and the task is still locomotion.
However, the direction the robot needs to move is switched every generation.
In the first generation, the objective of the robot is to move to the right, and in every next generation the direction objective alternates between left and right.
The robot does not have a sensor to know the direction it has to move in, the direction has to be learned during its lifetime.
An overview of this environment is shown in Figure \ref{fig:bidirectional-example}

The first plot of Figure \ref{fig:bidirectional} shows the results of this experiment.
Darwinian inheritance now shows to be beneficial, with a higher mean performance with both lifetime learning algorithms.
This is likely due to overfitting, where with Lamarckian inheritance, offspring inherit control parameters that are overfitted for the wrong movement direction.
This also explains the performance difference between \gls{bo} and \gls{rl}: The explorative nature of \gls{bo} can be used to find better control, despite the overfitted inherited control.

\subsection{Sensing environmental change}
\label{sec:exp-sensing}

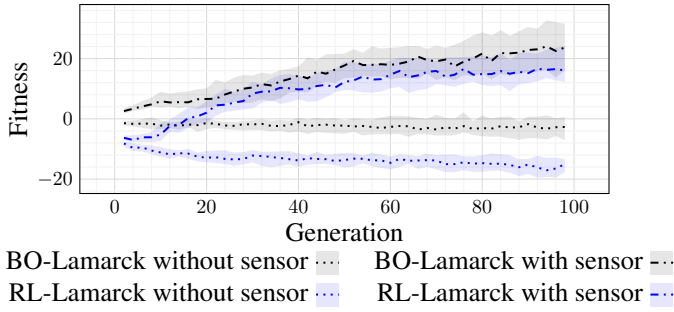
\begin{figure}
    \centering
    \begin{tikzpicture}[baseline=(current bounding box.center)]
        \begin{axis}[
            scale only axis,
            width=70mm,
            height=25mm,
            gridded,
            noinnerticks,
            xlabel={Generation},
            ylabel={Fitness}
        ]
            \lineminmax[lcolor=col.bo,ltype=linetype.nosensor]{results/before-learning/BObidirectional.txt}{}{x}{y}{ymin}{ymax}
            \lineminmax[lcolor=col.bo,ltype=linetype.sensor]{results/before-learning/BObidirectional2.txt}{}{x}{y}{ymin}{ymax}
            \lineminmax[lcolor=col.rl,ltype=linetype.nosensor]{results/before-learning/RLbidirectional.txt}{}{x}{y}{ymin}{ymax} 
            \lineminmax[lcolor=col.rl,ltype=linetype.sensor]{results/before-learning/RLbidirectional2.txt}{}{x}{y}{ymin}{ymax} 

        \end{axis}        
    \end{tikzpicture}    
    \begin{tabular}{rr}
        BO-Lamarck without sensor \addlegendimageintext{shaded legend image={col.bo}{linetype.nosensor}} &
        BO-Lamarck with sensor \addlegendimageintext{shaded legend image={col.bo}{linetype.sensor}} \\
        RL-Lamarck without sensor \addlegendimageintext{shaded legend image={col.rl}{linetype.nosensor}} &
        RL-Lamarck with sensor \addlegendimageintext{shaded legend image={col.rl}{linetype.sensor}}
    \end{tabular}
    \caption{
        Mean performance of the \emph{inherited} brain, \emph{i.e.} before learning, of the population over generations in a bidirectional environment, both with and without direction sensor. Shaded areas indicate interquartile range (25th to 75th percentile). The lines are smoothened with a window size of 2 generations.
    }
    \label{fig:before-learning}
\end{figure}

We showed that conflicting environmental changes can make Lamarckian inheritance less beneficial, as inherited learned behaviors may no longer match current conditions.
In nature, such conflicts are often mitigated by sensory information.
For example, foraging animals may inherit knowledge about which types of food are generally safe or nutritious, but the exact location of resources can change between generations.
If the animals can sense the presence of food from a distance, they can adjust their behavior to exploit current resources effectively, benefiting from inherited experience about types of food without being misled by outdated information about the location of food.
Inspired by this, we extended our bidirectional environment in robots by adding a direction sensor to the controller, providing the information necessary to exploit inherited learning effectively.
This is a single input to the \gls{ann} that indicates the goal direction.
The overall architecture, including number of hidden neurons and activation functions, of the brain remains the same.

The results of this final experiment are presented in the bottom plot of Figure \ref{fig:bidirectional}.
The plot clearly shows a benefit of Lamarckian inheritance over Darwinian inheritance.
When comparing it to the results of the previous section, we see that with Darwinian inheritance, there is no benefit of adding a direction sensor.
With Lamarckian inheritance, this sensor increases the performance significantly.
The greatest change can be seen with \gls{rl}: without a direction sensor, \gls{rl} with Lamarckian inheritance has the poorest performance.
However, with a direction sensor it has the best performance.
To visualize the effect of the direction sensor, we made videos of three example robots, where in the middle of the video the direction is switched \footnote{Download the videos here: \url{https://tinyurl.com/bde9jhy4}}.

% Anonymous: https://tinyurl.com/yc82uj23

\subsection{Sensor influence analysis}

\begin{figure}
    \centering
    \begin{tikzpicture}[baseline=(current bounding box.center)]
        \begin{groupplot}[
            group style={
                group size=1 by 2,
                xlabels at=edge bottom,
                ylabels at=edge left,
                horizontal sep=4mm,
                vertical sep=7mm
            },
            scale only axis,
            width=70mm,
            height=25mm,
            gridded,
            noinnerticks,
            xlabel={Robot-Controller pair (sorted by decreasing left movement)},
            ylabel={Fitness},
            xticklabels={,,},
            xtick=\empty,
        ]
        \nextgroupplot[title={Without sensor}]
            \lineminmax[lcolor=col.left,ltype=linetype.fixed,scatter]{results/fixed-robot/bidirectionalleft.txt}{}{x}{y}{ymin}{ymax}
            \lineminmax[lcolor=col.right,ltype=linetype.fixed,scatter]{results/fixed-robot/bidirectionalright.txt}{}{x}{y}{ymin}{ymax}
        \nextgroupplot[title={With sensor}]
            \lineminmax[lcolor=col.left,ltype=linetype.fixed,scatter]{results/fixed-robot/bidirectional2left.txt}{}{x}{y}{ymin}{ymax}
            \lineminmax[lcolor=col.right,ltype=linetype.fixed,scatter]{results/fixed-robot/bidirectional2right.txt}{}{x}{y}{ymin}{ymax}
        \end{groupplot}
    \end{tikzpicture}
    \begin{tabular}{rr}
        Left movement \addlegendimageintext{shaded legend image={col.left}{linetype.fixed}} &
        Right movement \addlegendimageintext{shaded legend image={col.right}{linetype.fixed}}
    \end{tabular}
    \caption{
        The performance of $10$ fixed robot morphologies with $10$ random sets of control parameters per morphology in the bidirectional environment settings. We evaluate the same morphology-control parameters pairs for both directions, where the difference is that in the second plot the robot has a direction sensor indicating the goal direction.
    }
    \label{fig:bidirectional-fixed}
\end{figure}
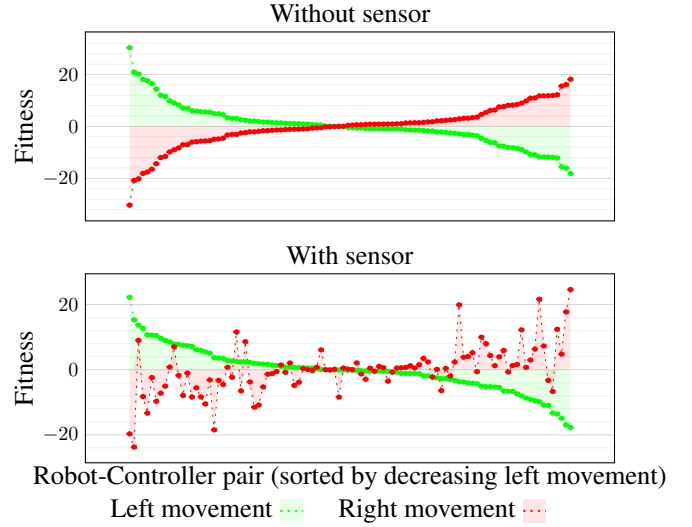

To see how well controller parameters transfer, we plotted the performance \emph{before} learning in Figure \ref{fig:before-learning}.
We define \emph{before learning} as the performance of the inherited control parameters.
Recall that in the \gls{bo} case, $8$ samples are inherited from parent to offspring, where in this case we select the best performing out of those samples.
We only show the values for the Lamarckian case, the performance of the inherited parameters for the robots in the Darwinian case was consistently around $0$.
The plot shows that, especially for \gls{rl}, the parameters do not transfer well without a direction sensor. 
With the added direction sensor, this is initially also the case, but after a few generations the inherited control parameters do transfer well.
Over generations, the robots learn to use the direction sensor to generalize locomotion, making it applicable for both directions.
These results also, again, show the strengths and weaknesses of the two learning methods.
\gls{bo} can quickly explore more of the search space, making it able to use the direction sensor more quickly.
However, the overall performance of \gls{rl} shown in the bottom plot of Figure \ref{fig:bidirectional} is still better.
This indicates an increasing and superior learning delta, showing the exploiting advantage of \gls{rl}.

Figure \ref{fig:bidirectional-fixed} shows the effect of adding a direction sensor to the robot controller.
We selected $10$ robot morphologies from the optimized set of robots, and evaluated $10$ \emph{random} sets of control parameters on the robots.
We again used the bidirectional environment, where we evaluated the exact same parameters for both left- and right-movement.
Without a direction sensor, the fitness for moving to the right is exactly the opposite of the fitness for moving to the left, which is as expected.
The movement will be exactly the same because there is no difference in input for the robot, and since the goal is reversed, the fitness will also be reversed.
With a direction sensor, the robot receives an additional input that changes depending on the direction.
With a decreasing fitness for moving to the left, the trend of moving to the right is still increasing, but there are some cases where the robot moves well to the left \emph{and} to the right.
These robots will have a higher chance of reproducing over multiple generations, because they have a consistent good performance.
The figure shows both the conflicting nature of the bidirectional movement, and how the direction sensor can be used to generalize movement in both directions.

\section{Conclusion}
When evolving robots, control parameters that have been optimized for one robot are potentially useful for other robots as well to not lose data over generations.
With Lamarckian inheritance, offspring robots inherit learned control parameters from their parent.
This is different from Darwinian inheritance, where the initial control parameters are inherited from parents to offspring.
Although earlier work beyond robotics report benefits of Darwinian inheritance over Lamarckian inheritance under dynamic environments, research in \gls{er} has shown that Lamarckian inheritance is beneficial in both stationary and dynamic environments.
We validate this finding in a dynamic rugged environment, where the benefit of Lamarckian inheritance remains consistent despite varying degrees of change.
We then show that, in dynamic settings, two variables influence the benefit of Lamarckian inheritance.
Firstly, it depends on how conflicting the environmental change is.
In our rugged environment, the change in environment did not seem to require conflicting control between environments.
In a bidirectional environment, where the goal direction changes every generation, we show a benefit of Darwinian inheritance when the environments are conflicting.
Secondly, it depends on the ability of the robot to predict the change of the environment.
When we give the robot a sensor that indicates the direction of the robot, over time it can use this information to generalize movement in both directions.
This will in turn benefit Lamarckian inheritance again.

Previous work, together with our results, suggests that Lamarckian inheritance remains beneficial for evolving robots in dynamic environments.
Even in scenarios where Darwinian inheritance is beneficial, we show that adding a simple sensor to detect environmental change causes the performance of Lamarckian inheritance to be sufficiently restored.
Future work could explore other conflicting environments, such as environmental changes that require morphological adaptations.
Additionaly, future work can investigate the potential of social learning, where robots inherit behaviors from peers in addition to direct ancestors.
Overall, our results show that the effectiveness of inheritance mechanisms is shaped by both environmental dynamics and the agent’s ability to predict them.

\balance

\section{Acknowledgements}
The work was supported by The Research Council of Norway (RCN) through its Center of Excellence scheme, RITMO with Project No. 262762, and through the Norwegian Center for Embodied AI (NCEI) under grant agreement no. 357451

\footnotesize
\bibliographystyle{apalike}
\bibliography{references}

\end{document}